%% file: main.tex
\documentclass[10pt]{article}

\usepackage[preprint]{rlj}

\usepackage{graphicx} 
\usepackage{array}
\usepackage{longtable}
\usepackage{subcaption}
\usepackage{booktabs}

\usepackage{amssymb}  

\usepackage{algorithm} 
\usepackage{algorithmic} 

\title{Black Box Meta-Learning Intrinsic Rewards}
\setrunningtitle{Black Box Meta-Learning Intrinsic Rewards}

\author{Octavio Pappalardo\textsuperscript{1},
Juan M. Santos\textsuperscript{2},
Rodrigo Ramele\textsuperscript{3}}

\emails{octaviopappalardo@gmail.com, \ 
juan.santos@unahur.edu.ar, \
rramele@itba.edu.ar}

\affiliations{
$^{1}$\textbf{Departamento de Física, Facultad de Ciencias Exactas y Naturales, Universidad de Buenos Aires }\\
$^{2}$\textbf{Centro de Investigacion en Informatica Aplicada, Universidad Nacional de Hurlingham}\\
$^{3}$\textbf{Departamento de Ingeniería Informática, Instituto Tecnológico de Buenos Aires}
}

\contribution{
We present a meta-reinforcement learning approach for learning components of a reinforcement learning algorithm. By treating the inner learning algorithm as a black box, this approach does not require the outer agent’s outputs to influence action selection in a differentiable manner. Additionally, it maintains memory and computational requirements independent of the inner adaptation method and enables simple outer updates that rely only on first-order gradients.
}
{
Within meta-reinforcement learning, a category of methods employs a reinforcement learning algorithm in the inner loop and learns components of this inner adaptation procedure that influence the resulting policy. Most methods in this category rely on meta-gradients computed by differentiating through the inner optimization process \citep{finn2017model,xu2018metagradient}. Alternative approaches with different trade-offs, such as first-order approximations \citep{finn2017model,nichol2018firstorder} and evolution strategies \citep{houthooft2018evolved,song2019maml}, have been explored. This paper examines optimizing the meta-objective with reinforcement learning while treating the inner policy optimization method as unknown.
    }
\contribution{
    Under the presented framework, we meta-learn an intrinsic reward function and an advantage function and validate their effectiveness.
    }
    { 
    We conduct a series of experiments in simulated robotic environments from MetaWorld \citep{yu2021metaworld}. In experiments with parametric task variations, we consistently observe higher task success rates when training PPO agents with the meta-learned intrinsic reward and advantage function compared to training them with hand-designed dense extrinsic rewards. The meta-learned components have access to the dense rewards during the meta-learning phase but rely solely on sparse success signals during testing. Our results focus on short adaptation phases of 4,000 steps and demonstrate the viability of the meta-learning approach.
    }


\summary{The broader application of reinforcement learning (RL) is limited by challenges including data efficiency, generalization capability, and ability to learn in sparse-reward environments. Meta-learning has emerged as a promising approach to address these issues by optimizing components of the learning algorithm to meet desired characteristics. Additionally, a different line of work has extensively studied the use of intrinsic rewards to enhance the exploration capabilities of algorithms. This work investigates how meta-learning can improve the training signal received by RL agents. We introduce a method to learn intrinsic rewards within a reinforcement learning framework that bypasses the typical use of meta-gradients by treating policy updates as unknown. We validate our approach against training with extrinsic rewards and additionally compare it to the use of a meta-learned advantage function.}

\begin{document}

\makeCover  
\maketitle  

\begin{abstract}
The broader application of reinforcement learning (RL) is limited by challenges including data efficiency, generalization capability, and ability to learn in sparse-reward environments. Meta-learning has emerged as a promising approach to address these issues by optimizing components of the learning algorithm to meet desired characteristics. Additionally, a different line of work has extensively studied the use of intrinsic rewards to enhance the exploration capabilities of algorithms. This work investigates how meta-learning can improve the training signal received by RL agents. We introduce a method to learn intrinsic rewards within a reinforcement learning framework that bypasses the typical computation of meta-gradients through an optimization process by treating policy updates as black boxes. We validate our approach against training with extrinsic rewards, demonstrating its effectiveness, and additionally compare it to the use of a meta-learned advantage function. Experiments are carried out on distributions of continuous control tasks with both parametric and non-parametric variations. Furthermore, only sparse rewards are used during evaluation. Code is available at: \url{https://github.com/Octavio-Pappalardo/Meta-learning-rewards}
\end{abstract}

\section{Introduction}\label{sec:Introduction}
\input{Introduction}

\section{Related Work}\label{sec:Related work}

\input{related_work}

\section{Black Box Meta-learning Intrinsic Rewards}\label{sec:Method_description}

\input{method_description}

\section{Experimental Setup}\label{sec:experimental methodology}
\input{exp_methodology}

\section{Experimental results}\label{sec:experimental results}
\input{exp_results}

\section{Conclusions}\label{sec:conclusions}
\input{conclusions}

\bibliographystyle{rlj}
\bibliography{main}

\beginSupplementaryMaterials

\input{supplementary_material}

\end{document}

%% file: Introduction.tex
The adoption of neural networks and recent algorithmic advances in reinforcement learning (RL) have enabled its application to complex decision-making problems \citep{schrittwieser2020mastering_MUZERO,degrave2022magnetic,luo2022controlling,mankowitz2023faster}. However, several challenges persist that must be addressed before RL can be effectively applied to a broader range of domains, especially to those requiring interaction with the real world.

Several sub-fields of research have emerged out of this need. Two main issues they target are poor data efficiency in learning tasks and the limited generalization capabilities of acquired policies when applied to new tasks. Related to the former, another central challenge in RL is devising efficient exploration strategies and achieving an appropriate trade-off with exploitation.

\textbf{Reinforcement Learning:} 
 In this work, we deal with discrete-time finite-horizon discounted Markov decision processes (MDPs). Reinforcement learning agents learn a policy $\pi_{\theta}$ that maps states to probability distributions over actions through interaction with an environment. The interaction takes place in episodes of length $T$. The state of the environment at step $t$ is $s_t \in S$, the action the agent takes is $a_t \in A $ and the reward it receives is $r_t$. The trajectory of an agent throughout an episode is denoted as $\tau=\left(s_0, a_0, r_1, \ldots, s_T\right)$ and the objective function it seeks to maximize is $J(\pi)=\mathbb{E}_{\tau \sim p(\tau)}\left[\sum_{t=1}^T \gamma^{t-1} r_t\right]=\mathbb{E}_{\tau \sim p(\tau)}[G(\tau)]$, where $\gamma$ is the MDP's discount factor and $G(\tau)$ is the discounted cumulative return attained in an episode of trajectory $\tau$.

\textbf{Meta Reinforcement Learning:} 
An agent's learning algorithm can be viewed as a mapping from the data $\mathcal{D}$ generated by interacting in an environment to the parameters of a policy $\pi_{\theta}$, $\theta=f(\mathcal{D})$.
The choice of learning algorithm $f$ is flexible and depends on a variety of decisions. If we parameterize the choice of a subset of this decisions with parameters $\phi$, then we can make the dependence explicit as $\theta=f(\mathcal{D} ; \phi)= f_{\phi}(\mathcal{D})$.
Meta-RL methods learn parameters $\phi$ such that the learning algorithm becomes more effective when applied to new tasks. 
A common setting for these methods involves a distribution of tasks $p(\mathcal{M})$ and an objective that drives meta-learning to maximize the expected cumulative return an agent attains throughout its lifetime when interacting with sampled tasks $\mathcal{M}^i \sim p(\mathcal{M})$ while learning with $f_{\phi}$. We define an agents \textit{lifetime} as its whole interaction with an environment (which spans multiple episodes). If the return attained at each episode of interaction with the environment is equally weighted by the objective, then it is described by Equation \ref{eq:obj_meta_RL}, but variations with different weightings are possible.

\begin{equation}
    \label{eq:obj_meta_RL}
    \mathcal{J}(\phi)=\mathbb{E}_{\mathcal{M}^i \sim p(\mathcal{M})}\left[\mathbb{E}_{\mathcal{D}}\left[\sum_{\tau \in \mathcal{D}} G(\tau) \; \Big| \; f_\phi, \mathcal{M}^i\right]\right].
\end{equation}

The meta-learning process operates at two different levels:
In the inner loop, given a task $\mathcal{M}^i$, the learning algorithm 
$f_{\phi}$ learns a task-specific policy $\pi_{\theta}$ through interaction with the task.
In the outer loop, a meta-learning algorithm learns parameters $\phi$ of $f$ using data from multiple inner loops.

\textbf{Intrinsic rewards:} 
For many RL algorithms, directly maximizing extrinsic rewards can lead to poor exploration of the environment; especially when dealing with sparse-reward environments. The use of intrinsic reward signals $r^i$, that either complement or replace extrinsic rewards $r=r^e$, has occupied a central role in searching for ways to achieve better exploration.

In this work, we draw from both these areas of research to meta-learn an intrinsic reward function. For this purpose, we model the reward function itself as a stochastic agent whose `actions' are the rewards it gives at each step and train it using an ordinary RL algorithm. This contrasts with the current standard of using meta-gradients\footnote{In this work, \textit{meta-gradients} refer specifically to gradients obtained by differentiating through the inner optimization process, rather than any estimate of the meta-objective's gradient with respect to the outer loop parameters.} when learning parts of the inner update. Our experiments evaluate the benefits of training a policy with these intrinsic rewards compared to using the environment's extrinsic signals. Additionally, we meta-learn an advantage function under the same framework and employ it for an alternative parameterization of the inner objective. All experiments are conducted on distributions of continuous control tasks, with access to shaped extrinsic rewards during meta-learning but only to sparse rewards during evaluation.

%% file: related_work.tex
One popular way to optimize for fast task learning is to use a neural network that uses all the data collected up to step $t$ to determine its action distribution at that step. During meta-training, the network is trained to leverage this interaction history to select appropriate actions for the current task. This family of methods was first introduced in \citet{duan2016rl} and \citet{wang2016learning}. 

Instead of meta-training a network to directly output the action taken at each step, another family of methods maintains the structure of standard reinforcement learning algorithms in the inner loop and meta-learns a component of this inner loop procedure that impacts the final performance. Most work restrict themselves to components that have a differentiable influence over the policy’s parameters and use meta-gradients for meta-training \citep{xu2018metagradient}. This framework was introduced in MAML \citep{finn2017model}, which meta-learns the initial parameters of a policy. Many other components have also been explored \citep{li2017meta,gupta2018meta,raghu2019rapid,park2019meta}. Some methods have avoided the use of meta-gradients by considering first-order approximations of them \citep{finn2017model,nichol2018firstorder}, using evolutionary algorithms \citep{salimans2017evolution,houthooft2018evolved,song2019maml}, or using a value based method \citep{sung2017_meta_critic}.

Several methods exist for designing intrinsic rewards \citep{oudeyer2007intrinsic,schmidhuber2010formal}, most of which are heuristic-based. Among these, popular approaches have been guided by the capacity to predict some aspect of the environment \citep{schmidhuber1991curious,pathak2017curiosity,burda2018exploration}, or have aimed to visit diverse states \citep{bellemare2016unifying,ostrovski2017count,hazan2019provably}. However, these are not the only possibilities. This work studies a different approach, it aims to \textit{learn} the intrinsic motivation signal \citep{optimal_rew_framework}.

Previous work has learned an intrinsic reward function using meta-gradients in a single task setting \citep{zheng2018learning,pmlr-v124-stadie20a,rajendran2020should} and using a distribution of tasks \citep{zheng2020can}. Additionally, \citet{alet2020meta} explored meta-learning a reward function with a discrete search over a space of programs. Moreover, rather than replacing extrinsic rewards, \citet{zou2019reward} learned to shape them into a denser signal. Among these methods, \citet{zheng2020can} has the closest  similarity to our work. The main differences are that they considered a deterministic reward function, trained it using meta-gradients, and conducted all their experiments in grid-world environments. Beyond the aforementioned work, several other methods have studied meta-learning parameterizations of the inner loop's objective function that don't utilize intrinsic rewards \citep{kirsch2019improving,zhou2020online,xu2020meta,oh2021discovering,bechtle2021meta,jackson2024discovering1,jackson2024discovering}.

%% file: method_description.tex
This work considers the inner loop learning algorithm $f$ to be a reinforcement learning method. Instead of using the standard reinforcement learning objective, in our approach $f$ uses an objective that is partly determined by a meta-learned neural network. This network is itself  trained with RL to maximize the meta-learning objective. Unlike previous methods that rely on meta-gradients to do so, our approach avoids the computation of second-order gradients. We achieve this by not modeling explicitly the influence intrinsic rewards have in the inner loop's learning procedure; instead, we let the outer loop treat it as part of the stochasticity involved in its optimization process. As a result, this approach doesn't require the ability to compute gradients of the policy's parameters with respect to the meta-learned parameters. Due to this considerations we refer to the method as being \textit{black box}.

We now describe in more detail how we meta-train the intrinsic reward function. We model the intrinsic reward function as a stochastic agent $\pi_{\phi}^{r}\left(r_t^i \mid \mathcal{D}_{:t}\right)$, where $\mathcal{D}_{:t}$ encodes all the interaction in an MDP up to time-step $t$. We use a LSTM \citep{hochreiter1997long} for this purpose. At each step $t$, the LSTM receives as input the tuple $\{ s_t, a_t , \pi_{\theta}(a_t \mid s_t) , r_t^e, r_{t-1}^i , \pi_{\phi}^{r}(r_{t-1}^i\mid \mathcal{D}_{:t-1}) , d_t \}$, where: $\pi_{\theta}$ is the policy that is being trained,  $r_t^e$ is the extrinsic reward received from the environment at time-step $t$ ,  $d_t \in \{0,1\} $ indicates whether $t$ is the start of a new episode,  $r_{t-1}^i$ is the intrinsic reward delivered at time-step $t-1$, and $\pi_{\phi}^{r}(r_{t-1}^i\mid \mathcal{D}_{:t-1})$ is the probability it was assigned by $\pi_{\phi}$. We train this network with PPO \citep{schulman2017proximal} to maximize a variant of the meta-learning objective defined in Equation \ref{eq:obj_meta_RL}, modified to add per episode discounting (\ref{sec:apend_hyp_explanations}). During meta-training, the network learns to use the history of interaction with the environment to generate outputs that aid the policy's learning process and lead to higher returns at the task it is currently facing. The resulting optimization scheme is similar to that of $RL^2$ \citep{duan2016rl,wang2016learning} but applied to learn a different quantity. The data for meta-training comes from running multiple inner loops that cover a distribution of training tasks. For a simplified and high-level overview of the proposed method, refer to the pseudocode in Algorithm \ref{code:intrinsic_rew}; further implementation details are discussed in later sections. We apply the same framework in section \ref{sec:exp_vs_adv} when learning an advantage function.

As mentioned, a distinguishing characteristic of our approach compared to previous methods is that it avoids the need to compute second-order gradients by bypassing the explicit modeling of the effect the meta-learned signal has over the policy's parameters. We now discuss some advantages and disadvantages of this black box approach versus the use of meta-gradients. The main potential drawback is that meta-gradients, by explicitly considering the role intrinsic rewards have in the inner loop, can provide a lower variance signal for meta-learning. This is discussed in \citet{stadie2019_E_MAML} in the context of meta-learning policy parameters. By comparison, the proposed method benefits from being simpler because it is framed as a standard reinforcement learning problem and doesn't require the calculation of gradients through an optimization process. Moreover, calculating second-order gradients is computationally more expensive than computing first-order gradients, making the black box approach favorable in this regard. Another related advantage is that the method is indifferent to how the inner loop uses the meta-learned component. In particular, the component can be employed in a non-differentiable manner to affect the choice of actions, whereas meta-gradient methods cannot be applied in such situations.

\begin{algorithm}[ht]
\caption{Meta-learning Intrinsic Rewards}
\begin{algorithmic}[1]\label{code:intrinsic_rew}
\STATE Initialize intrinsic reward agent $\pi_{\phi}^{r}$, outer loop critic, and task distribution $p(\mathcal{M})$
\WHILE{$\phi$ not converged}
    \STATE $\mathcal{D}_{\text{outer loop}} \gets \emptyset$
    \STATE $\text{batch\_of\_tasks} \gets \text{sample tasks from } p(\mathcal{M})$
    \FOR{each task $\mathcal{M}^i$ in $\text{batch\_of\_tasks}$}
        \STATE Initialize policy $\pi_{\theta}$ and inner loop critic
        \STATE $\mathcal{D}_{\text{life}} \gets \emptyset$
        \WHILE{lifetime not finished}
            \STATE $\mathcal{D}_{\text{inner loop}} \gets \text{collect data in } \mathcal{M}^i \text{ using } \pi_{\theta}$
            \STATE Update $\pi_{\theta}$ and inner loop critic with $\mathcal{D}_{\text{inner loop}}$ as in PPO, replacing environment rewards with intrinsic rewards from $\pi_{\phi}^{r}$
            \STATE $\mathcal{D}_{\text{life}} \gets \mathcal{D}_{\text{life}} \cup \mathcal{D}_{\text{inner loop}}$
        \ENDWHILE
        \STATE $\mathcal{D}_{\text{outer loop}} \gets \mathcal{D}_{\text{outer loop}} \cup \mathcal{D}_{\text{life}}$
    \ENDFOR
    \STATE Update $\pi_{\phi}^{r}$ and outer loop critic with $\mathcal{D}_{\text{outer loop}}$ as in PPO with respect to the meta-learning objective (see Equation \ref{eq:obj_meta_RL})
\ENDWHILE
\end{algorithmic}
\end{algorithm}

%% file: exp_methodology.tex
\subsection{Benchmarks}\label{sec:method_benchmarks}

We conduct all our experiments using the MetaWorld benchmarks \citep{yu2021metaworld}. These consist of distributions of continuous control tasks where a simulated robotic arm must interact with an object to achieve a desired configuration. Variations among tasks can be either non parametric (changes in the class of problem, e.g., opening a window, closing a drawer) or parametric (changes in the goal position or initial position within the same class of problem):

\begin{itemize}
\item ML1 benchmarks operate within a single problem class and contain 50 randomly sampled parametric variations for training and another 50 for evaluation.

\item ML10 benchmark consists of a set of 10 classes of problems for training and 5 different ones for evaluation. For each class, 50 sampled parametric variations are considered.
\end{itemize}

 All tasks share the same 39-dimensional observation space and 4-dimensional action space. Crucially, the agent lacks knowledge of both the task class and the goal position; it must infer these from interaction. All episodes are 500 steps long without the possibility of early termination (even if the task is completed).
Further details of the utilized benchmarks can be found in the supplementary material \ref{sec:apend_benchmarks}.

\textbf{Availability of Rewards:} \quad 
It is often easier and more practical to define desired behavior through sparse rewards (e.g., positive signal upon completion of an objective). However, the absence of frequent rewards makes learning significantly more challenging. This often leads to the manual design of a shaped reward function, which can be a labor-intensive process and increase the risk of reward hacking \citep{amodei2016concrete}. This work considers a hybrid setting in which there is access to shaped rewards for the training tasks but access to only sparse rewards during evaluation. In practice, this means we use shaped rewards in the meta-learning objective optimized in the outer loop, but we don't use them as inputs to the meta-learned recurrent network; here, only sparse rewards are considered. Prior work that operates in this same setting includes \citet{gupta2018meta,rakelly2019efficient,zhao2020meld}. The sparse rewards used are: $-0.2$ in the last step of episodes where the agent failed and $1 - 0{.}7 \frac{ \text{num. executed steps}} {\text{T}}$ on steps where the agent reaches the goal configuration (and has not done so previously within the episode).

\subsection{Implementation Details}
Two different neural network architectures were used in this work: an MLP for the policies and inner loop critics, and an LSTM for the meta-trained networks that output training signals and act as outer loop critics. A learning rate of $3 \times 10^{-4}$ was used in the inner loop and $5 \times 10^{-5}$ in the outer loop. We used the Adam optimizer, tanh activations, and orthogonal initialization. The outputs of the policies and the training signal networks are modeled as Gaussian distributions; the policies assume zero covariance.
During training, each lifetime used only 4,000 steps of data for updating the policy. Each outer loop update used data from 30 inner loops.
Further implementation details can be found in the supplementary material \ref{sec:apend_implementation_details} and in the study's code.

\subsection{Evaluation Methodology}\label{sec:eval_methodology}

During evaluation, the intrinsic reward function is used deterministically by considering the mean of the output Gaussian. In some experiments, making the policy deterministic after in-task training is also found to be beneficial.
The metric used for evaluation is the percentage of episodes in which the agent succeeded (i.e., reached a goal configuration).

The values reported for each algorithm in the results section are the average performance obtained from different instances of the method (each trained from scratch with a different seed) along with the corresponding standard deviation. For each benchmark, a given seed determines the set of parametric variations used. Five seeds were used for methods that utilize extrinsic rewards to train the policy, and three for those that use a meta-learned signal; larger numbers were not used due to the high computational demands of meta-RL algorithms and compute constraints.

Furthermore, for each seed, the method's performance on a given set of tasks (train or test set) was obtained as the average performance over multiple evaluation runs. This was done by running the method on each task of the set 10 independent times and averaging the performance across all these runs.
 For example, for the ML10 evaluation set, which has 5 different problem classes, each with 50 parametric variations, a total of 2,500 runs were considered to obtain the performance of each seed. Each run trains a randomly initialized policy on a task for 4,000 steps.

%% file: exp_results.tex
This section presents the main experimental results, divided into two parts. The first part compares training with meta-learned intrinsic rewards to training with extrinsic rewards. The second part investigates training a policy with a different parameterization of its loss that uses a meta-learned advantage function.

\subsection{Intrinsic vs Extrinsic Rewards}
This set of experiments evaluate whether there is benefit to be gained in making a reinforcement learning agent train with learned intrinsic rewards.
To this end, we compare the success rate of a PPO agent when trained with shaped extrinsic rewards, sparse extrinsic rewards, and intrinsic rewards generated by a meta-learned network (which only has access to the environment's sparse rewards).

\begin{figure}[ht]
  \centering
  \includegraphics[width=1.0\columnwidth]{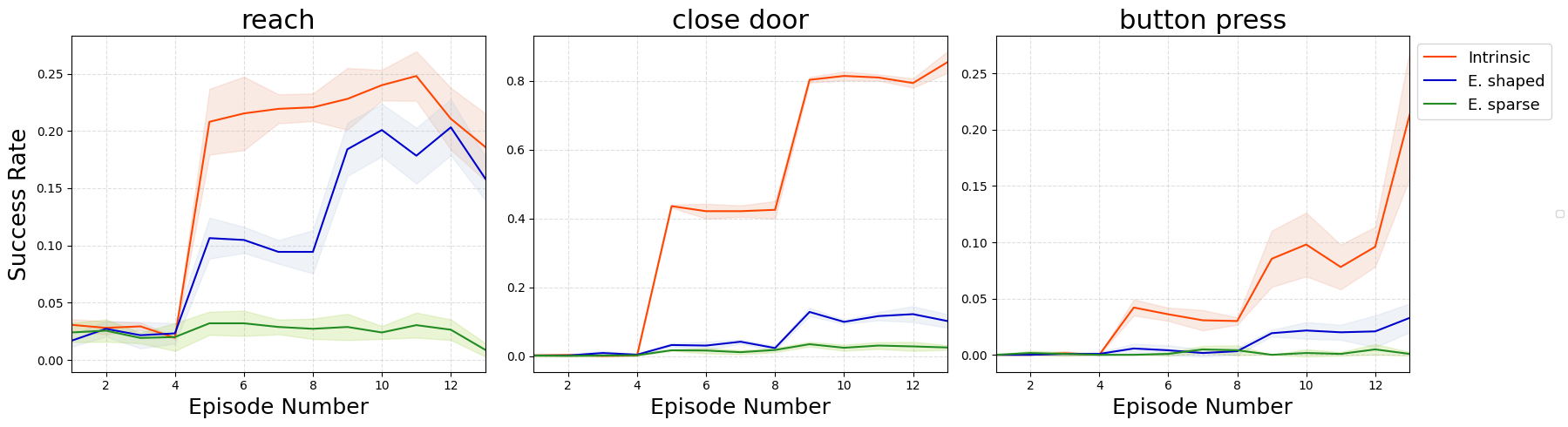}
  \caption{Comparison of the average performance of agents as they interact with tasks from the test set. The values and their standard deviations (represented by the shaded region) were obtained as explained in section \ref{sec:eval_methodology}. The success rate when training agents using three different types of rewards is shown: intrinsic (red), shaped extrinsic (blue), and sparse extrinsic(green). Three benchmarks are considered: ML1-reach, ML1-close-door, and ML1-button-press. The last episode reflects the performance of the final policy when made deterministic.}
  \label{fig:contra_extr_curva}
\end{figure}

\begin{figure}[ht]
  \centering
  \includegraphics[width=0.8\columnwidth]{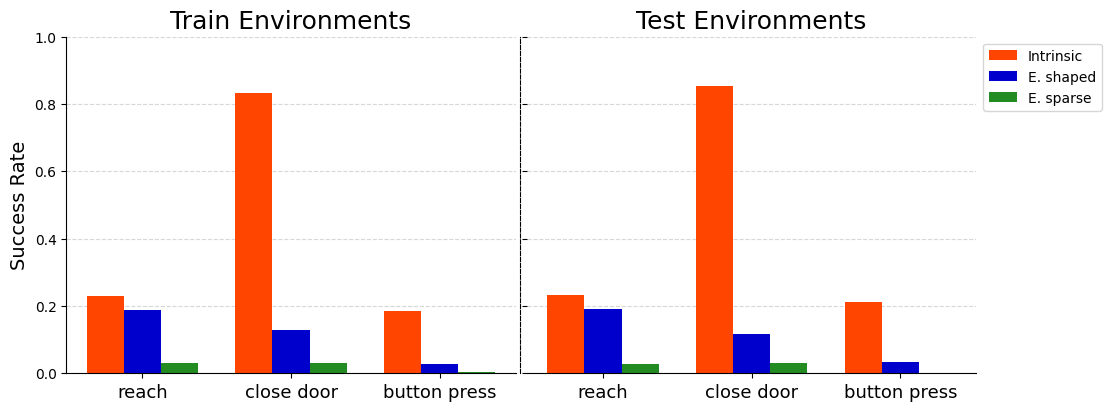} 
  \caption{Success rate of agents trained with different rewards on various meta-learning benchmarks, ML1-reach, ML1-close-door, and ML1-button-press, after an adaptation period of 4,000 steps. The figure compares the performance when using three different types of rewards: intrinsic (red), shaped extrinsic (blue), and sparse extrinsic(green). The values were obtained as explained in section \ref{sec:eval_methodology}.}
  \label{fig:contra_extr_bar}
\end{figure}

Evaluations are conducted over the ML1-reach, ML1-close-door, and ML1-button-press benchmarks, allowing only 4,000 steps of training. Figures \ref{fig:contra_extr_curva} and \ref{fig:contra_extr_bar} illustrate the results.
Both figures demonstrate substantial improvements when training with an intrinsic reward function. Figure \ref{fig:contra_extr_curva} evidences the choice of using an RL algorithm that trains with batches of 4 episodes of data (\ref{sec:apend_hyp_tables}). It also reflects that making the policy deterministic after training can improve performance in some environments. Figure \ref{fig:contra_extr_bar} shows that there is no decline in performance when moving from tasks in the training set to tasks in the test set. While this is expected when training with extrinsic rewards (since no meta-learned component is applied and all tasks are sampled from the same distribution), the fact that the same behavior occurs when training with intrinsic rewards indicates that the learned reward network effectively generalizes to  unseen environments within the distribution.

As mentioned, training with intrinsic rewards showed significant improvements in performance compared to using either type of extrinsic rewards. It is worth noting that in this evaluation setting the meta-learned network only has access to the sparse extrinsic rewards, which makes the gains more notable. Training directly with sparse rewards showed little to no progress. Moreover, the fact it also outperforms training with shaped extrinsic rewards suggests learning rewards can also be useful for the design or improvement of reward signals used in standard RL benchmarks \citep{zou2019reward}. However, the observed benefits do not come without costs. The method using intrinsic rewards leverages a prior meta-learning phase which can be computationally costly and requires having access to training tasks that share structure with tasks in the test set.

\subsection{Intrinsic Rewards vs Learned Advantages}\label{sec:exp_vs_adv}

In this section, we consider whether intrinsic rewards are the right component to meta-learn and we discuss some other options. Focusing on intrinsic rewards is appealing for at least two reasons: 1) as mentioned in section \ref{sec:Related work}, the use of intrinsic rewards is a well-studied topic in RL and has consistently shown benefits, 2) all standard RL algorithms assume they receive rewards; thus, meta-learned rewards can be directly integrated into any of them while maintaining their structure.

\begin{figure}[ht]
    \centering
    \begin{subfigure}[b]{1.0\columnwidth}
        \centering
        \includegraphics[width=1.0\columnwidth]{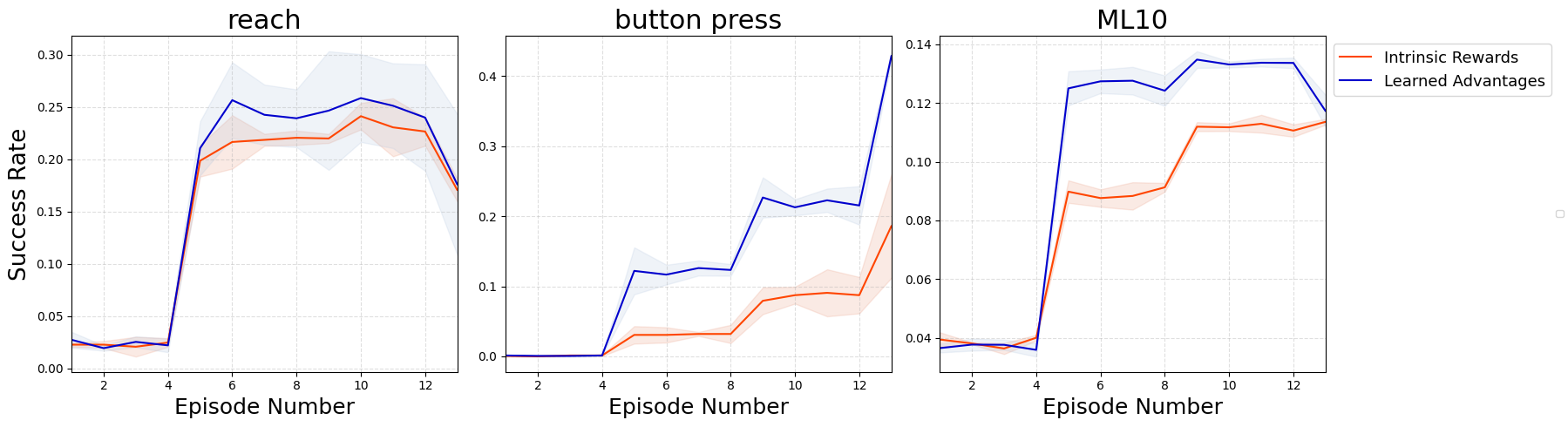}
        \caption{Train environments}
        \vspace{12pt}
    \end{subfigure}
    
    \begin{subfigure}[b]{1.0\columnwidth}
        \centering
        \includegraphics[width=1.0\columnwidth]{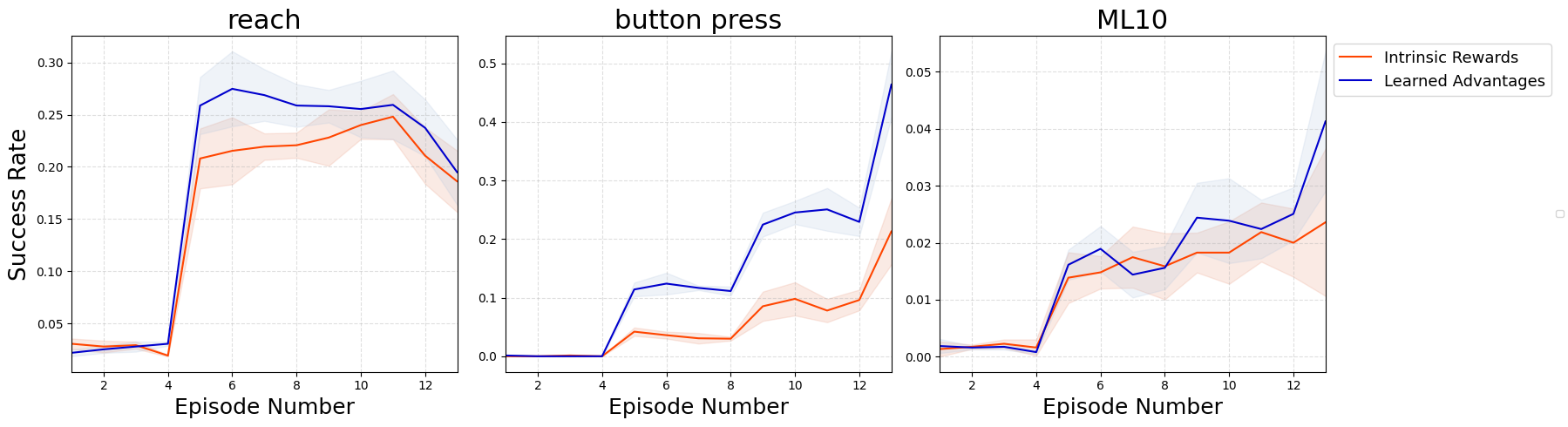}
        \caption{Test environments}
    \end{subfigure}
    \caption{Comparison of the average performance of meta-learning methods as they interact with a new task from the training set (a) and the testing set (b). Success rates are shown for two methods that meta-learned different parameterizations of the loss: using intrinsic rewards (red) and using advantages (blue). Three benchmarks are considered: ML1-reach, ML1-button-press, and ML10. The last episode reflects the performance of the final policy when made deterministic.}
    \label{fig:contra_ventaja_curvas}
\end{figure}

As discussed in section \ref{sec:Related work}, it is possible to meta-learn several other parameterizations of the reinforcement learning objective. In this section, we explore  another such parameterization by meta-learning an advantage function. Instead of learning to assign partial credit to a transition and its preceding transitions, the network learns to evaluate each transition's quality independently. This set of experiments also introduces non-parametric variations among environments. Results are shown in figure \ref{fig:contra_ventaja_curvas}.

Both methods exhibit similar qualitative behavior. For the considered benchmarks, using the learned advantage function shows some benefits. These improvements are statistically significant only for ML1-button-press and for the training tasks of ML10. While both methods show good generalization when facing parametric variations, they are unable to maintain their performance when dealing with classes of tasks that differ from those seen during training. Even though updates with learned intrinsic rewards and advantages still manage to improve randomly initialized policies in the test tasks of ML10, they lead to considerably lower success rates than those obtained in the training tasks.

We clarify that we refer to the meta-learned component in this section as an advantage function because of the role its outputs play when computing PPO updates in the inner loop. Compared to meta-learning rewards, this approach disregards more of the inner algorithm's structure and is closer to prior work that meta-learned the entire inner objective. A meta-learned value function can also be used to directly choose actions; the same meta-learning procedure would work without requiring any changes because the outer updates make no assumptions about how the network's outputs are used in the inner loop. In our work, similar flexibility can be seen in the choice of the inner loop algorithm. While meta-gradient methods typically use simpler vanilla policy gradient updates in the inner loop, we employ PPO updates without any added complexity or computational cost to the outer loop.

Meta-learning to predict the training signal can be complementary to learning other components of the inner loop. Training signals hold knowledge on the goodness of observed behaviors \citep{zheng2020can}. Other components can address different loci of knowledge and provide additional benefits. In particular, the most popular approach in meta-learning literature is to meta-learn parameters of a policy. 
One advantage of meta-learning part of the objective function is its applicability to broad distributions of environments and agents. It has also been shown to be adequate for the many-shot meta-RL setting \citep{kirsch2019improving,oh2021discovering}.
Determining the optimal combination of components to meta-learn remains an open problem and is likely dependent on the specific setting.

%% file: conclusions.tex
This paper meta-learned a recurrent and stochastic intrinsic reward function in a manner analogous to how standard reinforcement learning agents are trained. This approach presents an alternative to using meta-gradients and evolution strategies. The network was trained via interaction with a distribution of training tasks and later evaluated in unseen environments with sparse rewards. Our experiments demonstrate that training a policy using the learned reward function can significantly accelerate learning compared to using extrinsic rewards. The intrinsic reward function demonstrated effective generalization across parametric task variations, but it struggled when applied to problem types not seen during meta-learning. Additionally, we also meta-learned an advantage function under the same framework and achieved slightly better results when using it to train a policy. Throughout the paper, we discussed the benefits and drawbacks of different features of our approach.

Research directions that can stem from this work to provide further insights into the proposed approach include extending the method to longer lifetimes and broader task distributions and conducting a quantitative analysis comparing this black box approach to the use of meta-gradients or evolution strategies. Moreover, this work considered sparse rewards only during evaluation; it would be beneficial to progress towards settings where only sparse rewards are available during meta-learning as well. Some straightforward avenues that can lead to better sample efficiency are to learn initial parameters for the policy instead of using random initialization and to use a network that has access to future interaction steps within the batch of collected data when generating rewards, rather than a network that only looks at past data.

%% file: supplementary_material.tex
{
\setcounter{section}{0}
\renewcommand{\thesection}{S.\arabic{section}} 
\renewcommand{\thesubsection}{\thesection.\arabic{subsection}} 

Section \ref{sec:apend_benchmarks} gives more information on the MetaWorld benchmarks involved in the study. Section \ref{sec:apend_implementation_details} provides more details on the algorithms' implementations.

\section{Benchmarks}\label{sec:apend_benchmarks}
This section gives more information on the utilized benchmarks \citep{yu2021metaworld}. They all are from version 2 of MetaWorld.

\vspace{6pt}
\textbf{Observation and action space:}
The 4-dimensional action space contains 3 dimensions for the change in 3D space of the robotic arm's end-effector plus 1 dimension for the gripper's normalized torque. All dimensions are bounded to the range $[-1,1]$. The 39-dimensional observation space consists of the 3D Cartesian position of the end-effector, a normalized measurement of how
open the gripper is, the 3D position and quaternion of the first and second object of interest, all this same information from the previous time step, and finally, the 3D position of the goal. In the benchmarks we utilized, the  3D position of the goal is zeroed out which forces the agent to learn to recognize and adapt to the task through interaction. For tasks where there isn't a second object of interest this information is also zeroed out.

\vspace{6pt}
\textbf{ML1 benchmarks:}
The choice of these benchmarks over others in the ML1 category was arbitrary except for the condition that the agent trained with shaped extrinsic rewards exhibited some progress during its lifetime.
\begin{itemize}
    \item \textbf{`ML1 reach'} Reach a goal position. Randomize the goal positions.
    \item \textbf{`ML1 close door'} Close a door with a revolving joint. Randomize door positions.
    \item \textbf{`ML1 button press'}  Press a button. Randomize button positions
\end{itemize}

\vspace{6pt}
\noindent{\textbf{ML10 benchmark:}}

\textbf{Training environments}

\begin{itemize}
    \item insert peg side: Insert a peg sideways. Randomize peg and goal positions.
    \item reach-v2: Reach a goal position. Randomize the goal positions.
    \item door-open-v2: Open a door with a revolving joint. Randomize door positions.
    \item basketball-v2: Dunk the basketball into the basket. Randomize basketball and basket positions.
    \item open window: Push and open a window. Randomize window positions.
    \item pick \& place: Pick and place a puck to a goal. Randomize puck and goal positions.
    \item button-press-topdown-v2: Press a button from the top. Randomize button positions.
    \item push-v2: Push the puck to a goal. Randomize puck and goal positions.
    \item close drawer: Push and close a drawer. Randomize the drawer positions.
    \item sweep-v2: Sweep a puck off the table. Randomize puck positions.
\end{itemize}
\textbf{Test environments}
\begin{itemize}
    \item door-close-v2: Close a door with a revolving joint. Randomize door positions.
    \item place onto shelf: Pick and place a puck onto a shelf. Randomize puck and shelf positions.
    \item drawer-open-v2: Open a drawer. Randomize drawer positions.
    \item sweep into hole: Sweep a puck into a hole. Randomize puck positions.
    \item pull lever: Pull a lever down 90 degrees. Randomize lever positions.
\end{itemize}


\section{Implementation Details}\label{sec:apend_implementation_details}

Section \ref{sec:apend_hyp_tables} presents tables with the values of different hyperparameters for each method. Their nomenclature matches that used in the project's code for easier reproducibility. Then, in section \ref{sec:apend_hyp_explanations} we give a discussion on the influence of various hyperparameters, and provide some additional information.

\subsection{Hyperparameter Tables}\label{sec:apend_hyp_tables}

\textbf{PPO agent/ Inner loop}

Hyperparameters for the inner loop PPO agent. The hyperparameters are the same for all variations of training signals: extrinsic shaped rewards, extrinsic sparse rewards, intrinsic learned rewards, and learned advantages.

\vspace{2pt}

\begin{longtable}{|p{4.5cm}|p{4cm}|}
\hline
\textbf{Hyperparameter} & \textbf{Value} \\
\hline
\endfirsthead
\hline
\textbf{Hyperparameter} & \textbf{Value} \\
\hline
\endhead
\hline
\endfoot
\hline
\multicolumn{2}{r}{} \\[-12pt] 
\caption{Table with the hyperparameter configuration used for task adaptation. Valid both for when the inner PPO agent was runned on its own. as well as when it used meta-learned training signals. hyperparameters were also the same during meta-training phases.} \label{tab:PPO_inner_hiperparametes}
\endlastfoot

\texttt{total\_timesteps} & \texttt{6,000} \\
\hline
\texttt{num\_steps} & \texttt{2,000} \\
\hline
\texttt{learning\_rate} & \texttt{3e-4} \\
\hline
\texttt{adam\_eps} & \texttt{1e-5} \\
\hline
\texttt{gamma} & \texttt{0.99} \\
\hline
\texttt{gae} & \texttt{True} \\
\hline
\texttt{gae\_lambda} & \texttt{0.95} \\
\hline
\texttt{ppo.update\_epochs} & \texttt{64} \\
\hline
\texttt{ppo.num\_minibatches} & \texttt{16} \\
\hline
\texttt{ppo.normalize\_advantage} & \texttt{True} \\
\hline
\texttt{ppo.clip\_coef} & \texttt{0.2} \\
\hline
\texttt{ppo.entropy\_coef} & \texttt{0.0} \\
\hline
\texttt{ppo.valuef\_coef} & \texttt{0.5} \\
\hline
\texttt{ppo.clip\_grad\_norm} & \texttt{True} \\
\hline
\texttt{ppo.max\_grad\_norm} & \texttt{0.5} \\
\hline
\texttt{ppo.target\_KL} & \texttt{None} \\
\hline 
\end{longtable}

\textbf{Learned Intrinsic rewards and Advantage function/ Outer loop}

\vspace{2pt}

\begin{longtable}{|p{6cm}|p{4.5cm}|}
\hline
\textbf{Hyperparameter} & \textbf{Value} \\
\hline
\endfirsthead
\textbf{Hyperparameter} & \textbf{Value} \\
\hline
\endhead
\hline
\endfoot
\hline
\multicolumn{2}{r}{} \\[-12pt] 
\caption{Hyperparameter configuration for training and evaluating the meta-agents that output intrinsic rewards and advantages.} \label{tab:hperparams_intr_and_adv} 
\endlastfoot

\texttt{num\_epsiodes\_of\_validation} & \texttt{4} \\
\hline
\texttt{num\_lifetimes\_for\_validation} & \texttt{60} \\
\hline
\texttt{num\_inner\_loops\_per\_update} & \texttt{30} \\
\hline
\texttt{learning\_rate} & \texttt{5e-5} \\
\hline
\texttt{adam\_eps} & \texttt{1e-5} \\
\hline
\texttt{e\_rewards\_target\_mean} & \texttt{0.0001} \\
\hline
\texttt{meta\_gamma} & \texttt{0.9} \\
\hline
\texttt{ae.estimation\_method} & \texttt{`bootstrapping skipping uninfluenced future rewards'} \\
\hline
\texttt{ae.bootstrapping\_lambda} & \texttt{0.85} \\
\hline
\texttt{ae.starting\_n} & \texttt{2,200} \\
\hline
\texttt{ae.num\_n\_step\_estimates} & \texttt{6} \\
\hline
\texttt{ae.skip\_rate} & \texttt{300} \\
\hline
\texttt{rnn\_input\_size} & \texttt{32} \\
\hline
\texttt{rnn\_type} & \texttt{lstm} \\
\hline
\texttt{rnn\_hidden\_state\_size} & \texttt{128} \\
\hline
\texttt{initial\_std} & Intrinsic r.=\texttt{0.2} \newline Advantages=\texttt{1.0}  \\
\hline
\texttt{ppo.k} & \texttt{400} \\
\hline
\texttt{ppo.update\_epochs} & \texttt{12} \\
\hline
\texttt{ppo.num\_minibatches} & $"0" \approx \left( \frac{\texttt{num inner loop steps}}{ \texttt{ppo.k}} \right) $ \\
\hline
\texttt{ppo.normalize\_advantage} & \texttt{True} \\
\hline
\texttt{ppo.clip\_coef} & \texttt{0.2} \\
\hline
\texttt{ppo.entropy\_coef} &  \texttt{0.0005  \ }  \texttt{Intr.r.ML1=0.003} \\
\hline
\texttt{ppo.valuef\_coef} & \texttt{0.5} \\
\hline
\texttt{ppo.clip\_grad\_norm} & \texttt{True} \\
\hline
\texttt{ppo.max\_grad\_norm} & \texttt{0.5} \\
\hline
\texttt{ppo.target\_KL} & \texttt{0.01} \\
\hline
\end{longtable}

\subsection{Hyperparameter Explanations and Further Implementation Details}\label{sec:apend_hyp_explanations}

\textbf{Architectures:}
For the policy and critic we used a (64,64) MLP architecture. For the meta-learned networks we used an LSTM with a hidden dimension of 128. Prior to going into the LSTM, the current step's input is processed by two linear (plus activation) layers (128,32), at which all 1 dimensional inputs were concatenated. The LSTM hidden states were processed by a (512) layer for the outer loop critic, a (128) layer for the gaussian distribution standard deviation, and (128,128) layers with a final arctan activation for its mean. All other activations were tanh. When using the learned advantage function the critic was omitted from the inner loop. All algorithms where implemented in PyTorch.

\textbf{Hyperparameter details:}
Several of the hyperparameters showed in section \ref{sec:apend_hyp_tables} are standard in deep RL literature. This section briefly describes those whose influence may be less clear.

\begin{itemize}

 \item  \textit{total\_timesteps} refers to the total number of steps collected in inner loops - i.e. the length of lifetimes.  \textit{num\_steps} is the batch size used for PPO updates. Two updates are executed during an inner loop.
 
 \item \textit{num\_epsiodes\_of\_validation} and \textit{num\_lifetimes\_for\_validation} control how many of the last lifetimes and episodes whithin them are used to validate performance during training. Meta-learning took place until no appreciable performance improvement was observed for at least 200 outer loop updates.

\item Throughout meta-training, extrinsic rewards obtained in each inner loop were normalized to keep their average value around \textit{e\_rewards\_target\_mean}. This normalization was done at the level of each environment class. In the ML10 benchmark training, independent averages of extrinsic rewards for each of the 10 problem classes were maintained. This normalization ensures that the meta agent's training signals (extrinsic rewards) stay within the same range during training and assigns similar importance to improvements across different environment classes.

\item Hyperparameters starting with \textit{ae} and \textit{meta\_gamma} control which variant of Objective \ref{eq:obj_meta_RL} is used and how advantages are estimated for outer loop updates. The most conceptually important are \textit{ae.estimation\_method} and \textit{meta\_gamma}. \textit{ae.estimation\_method} controls whether discounts are applied at the episode or step level and which method is used for estimating advantages. Specifically, \textit{`bootstrapping skipping uninfluenced future rewards’} uses episode-wise discounts and estimates advantages with an exponential combination of n-step estimates. This method ignores all extrinsic rewards received by the agent until the generated training signal is used for an update. \textit{meta\_gamma} indicates the value of the discount factor used for outer loop learning (no matter if this is for step-wise or episodic discounting). 

\item \textit{initial\_std} controls the initial standard deviation of the generated training signals when meta-training.

\item Hyperparameters starting with \textit{ppo} for meta-learning have similar but not identical roles to the standard PPO  hyperparameters. PPO updates for the outer loops use truncated backpropagation. Specifically, \textit{ppo.k} controls the number of steps before gradient propagation is truncated. \textit{ppo.num\_minibatches} controls the number of processed k-length sequences before a gradient step is taken. The value "0" in the tables indicates a gradient step is taken for each num\_inner\_loops k-length sequences.

\item Hyperparameters whose values were not mentioned in the tables took the default value assigned in Pytorch.
\end{itemize}

Some of the values that we found to impact performance the most are: \textit{initial\_std}, \textit{entropy\_coef }, \textit{e\_rewards\_target\_mean}, \textit{num\_inner\_loops\_per\_update } and learning rates. Ray was used to run the different inner loops corresponding to an outer loop update in parallel across different processes.


}